\documentclass{article}
\usepackage{amsmath}
\usepackage{bm}
\usepackage{amssymb}
\usepackage{graphicx}
\usepackage[title]{appendix}
\usepackage{gensymb}
\title{3DOF+Quantization: 3DGS quantization for large scenes with limited Degrees of Freedom}
\author{Matthieu GENDRIN  \\ Stéphane PATEUX \\ Théo LADUNE \\
	Orange Innovation
	}
\date{\today}
\begin{document}

\maketitle

\begin{abstract}
3DGS \cite{kerbl20233d} is a major breakthrough in 3D scene reconstruction.
With a number of views of a given object or scene, the algorithm trains a model 
composed of 3D gaussians, which enables the production of novel views from arbitrary
points of view. This freedom of movement is referred to as 6DoF for 6 degrees of freedom:
a view is produced for any position (3 degrees), orientation of camera (3 other degrees).
On large scenes, though, the input views are acquired from a limited zone in space,
and the reconstruction is valuable for novel views from the same zone, even if the scene 
itself is almost unlimited in size.
We refer to this particular case as 3DoF+, meaning that the 3 degrees of freedom of camera 
position are limited to small offsets around the central position. 
Considering the problem of coordinate quantization, the impact of position
error on the projection error in pixels is studied.
It is shown that the projection error is proportional to the squared inverse distance of the point being projected.
Consequently, a new quantization scheme based on spherical coordinates is proposed.
Rate-distortion performance of the proposed method are illustrated on the well-known Garden scene.
\end{abstract}

\section{Introduction}
3DGS \cite{kerbl20233d} has opened new possibilities in terms of novel view synthesis of 3D scenes.
The quality and training performance are such that a major part of the 3D research community has
switched to this model.
With this success comes the need to compress such models, which is achieved in several ways in the literature.
Papantonakis\textit{ et al.} \cite{papantonakis2024reducing} proposes to optimize the number of gaussians and the color coefficients,
and to use a codebook-based quantization method.
Scaffold-GS (\cite{lu2024scaffold}) introduces a structured description
of the model to obtain additional compression performance. HAC (\cite{chen2024hac}) builds upon Scaffold-GS adding entropy minimization for further rate savings.
These previous work provides compelling rate-distortion, without any hypothesis on the
degrees of freedom of the camera. 
As a complement, this paper analyses how the 3DoF+ hypothesis can be leveraged
to perform a more accurate bit allocation to the spatial coordinates, giving more precision to the gaussians near the cameras, to the
expense of gaussians located further away. Note that this work is complementary to the existing methods.

\section{Preliminaries}

3D Gaussian Splatting (3DGS) \cite{kerbl20233d} models a 3D scene with 3D gaussians,
and renders viewpoints through a differentiable splatting and tile-based rasterization. 
Each Gaussian is defined by a 3D covariance matrix $\bm{\Sigma} \in \mathbb{R}^{3\times3}$ and location (mean) $\bm{\mu} \in \mathbb{R}^{3}$,
where $\mathbf{x} \in \mathbb{R}^{3}$ is a random 3D point, and $\bm{\Sigma}$ is defined by a diagonal matrix
$\mathbf{S} \in \mathbb{R}^{3\times3}$ representing scaling and rotation matrix $\bm{R_g} \in \mathbb{R}^{3\times3}$ to guarantee its
positive semi-definite characteristics, such that $\bm{\Sigma} = \bm{R_g}\bm{S}\bm{S}^{\top}\bm{R_g}^{\top}$.

\begin{align}
    \label{exp}
    G(\mathbf{x}) = exp(- \frac{1}{2} (\mathbf{x} - \bm{\mu})^{\top} \bm{\Sigma}^{-1} (\mathbf{x} - \bm{\mu}))
\end{align}

\begin{figure}
    \includegraphics[width=\linewidth]{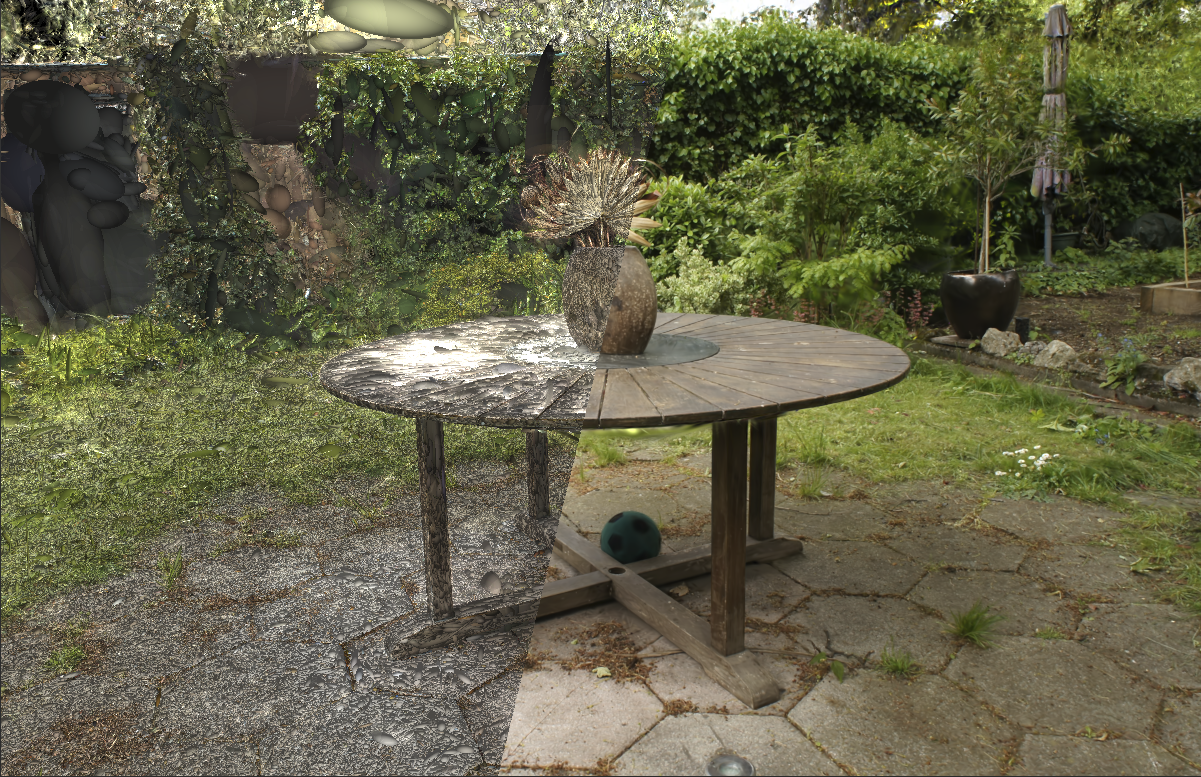}
    \caption{Gaussian render example.
    The rendering has been modified on the left part to highlight the gaussians,
    while the right part is the standard 3DGS rendering.}
    \label{fig:3dgs}
\end{figure}

To render an image from a random viewpoint, 3D Gaussians are first splatted to 2D, and
render the pixel value $\mathbf{C} \in \mathbb{R}^{3}$ using $\alpha$-composed blending,
where $\alpha \in \mathbb{R}$ measures the contribution to this pixel of each Gaussian after 2D projection, $\mathbf{c} \in \mathbb{R}^{3}$
is view-dependent color modeled by Spherical Harmonic (SH) coefficients, and
$N$ is the number of sorted Gaussians contributing to the rendering.
The 3DGS rendering is illustrated in figure \ref{fig:3dgs}.

\begin{align}
    \mathbf{C} = \sum_{n \in N} c_n \alpha_n \prod_{j = 1}^{n - 1} (1 - \alpha_j)
\end{align}

Note that the gaussian calculation exposed in eq. \eqref{exp} is actually done in 
2D after projection of the gaussians center and covariance matrix on the screen image.
The projection of the center is done with the classical formulation given in eq. \eqref{persp}.

\section{Position dependent quantization methodology}

\paragraph{Perspective projection}

For the sake of simplicity, let us first consider a toy system with a single
camera. Any point with coordinates $(x,y,z)$ in the camera referential is projected to the image plan 
as $(u,v)$:
\begin{align}
\label{persp}
    u = f \frac{x}{z} \\
    v = f \frac{y}{z} \nonumber
\end{align}

Under a high-rate hypothesis, quantizing the point coordinates $(x,y,z)$ with a
scalar, unitary quantizer can be modeled as adding an independent noise $\delta$ to the coordinates.
\begin{enumerate}
    \item the impact on $(u,v)$ of the noise $\delta$ on $x,y$ is proportional to $\frac{1}{z}$
    \item the impact on $(u,v)$ of the noise $\delta$ on $z$ is proportional to $\frac{1}{z^2}$
\end{enumerate}

Of course, we don't want to encode the model for one precise camera, but 
This gives the intuition of how the 3DoF+ assumption can be leveraged: the 
$z$ coordinate of the local referential is to be quantized differently than the $(x,y)$.
For example, we could use larger quantization steps for $x,y,z$ when $z$ is large.
And even larger step for $z$ than for $x,y$.
\footnote{Note that this encourages to use spherical coordinates here for performing
positional quantization. Simple scalar quantization on spherical coordinates would
then lead to well adapted vector quantization of positions.}
To generalize from this toy example, we note that $z$ is approximately the distance between the camera and the gaussian, and that the
cameras are all in the same area in 3DoF+. As such, the reasoning holds for all cameras.

\begin{figure}
    \begin{center}
        \includegraphics[height=5cm]{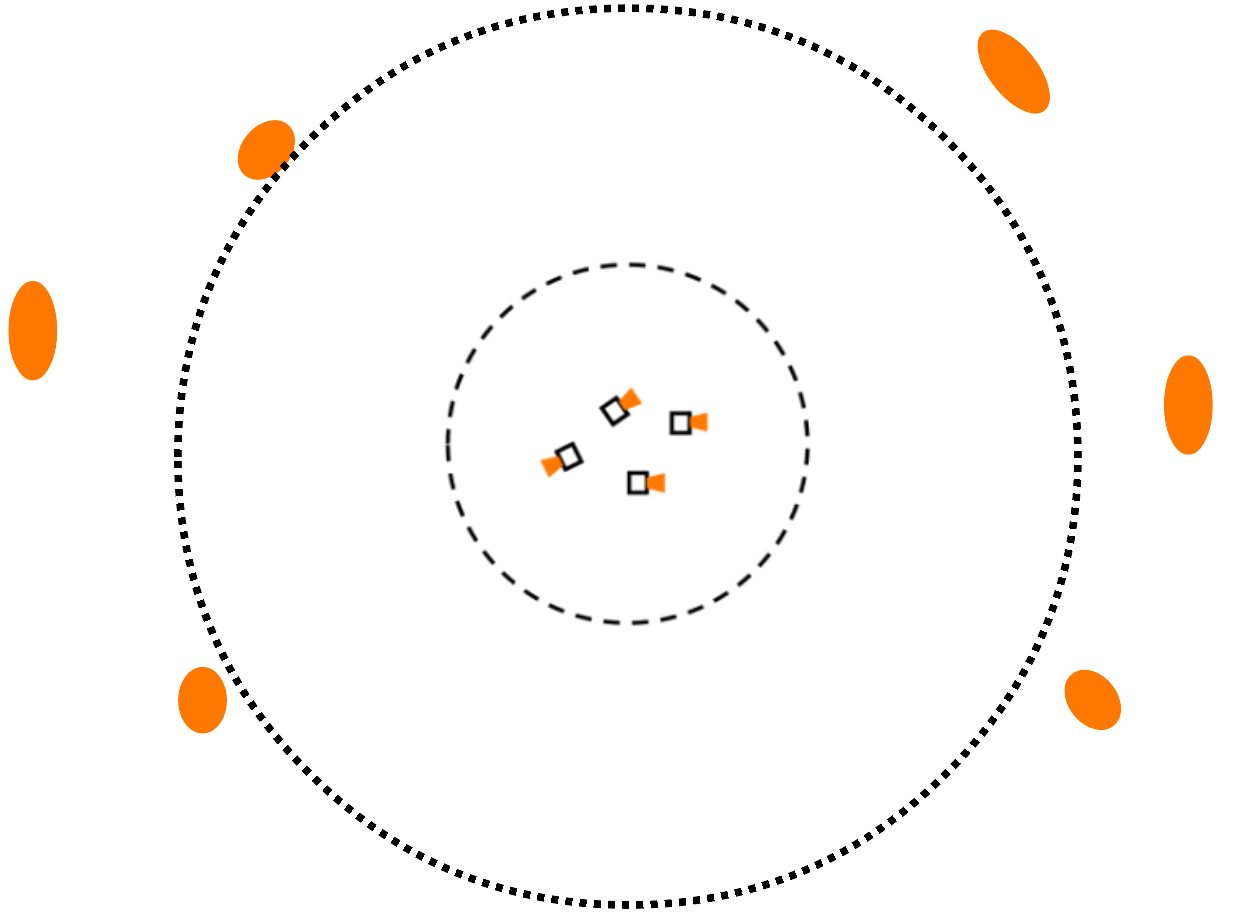}
        \caption{3DoF+ scene. \\
        Inner circle, radius $R_i$, is the limit of the possible camera poses. \\
        Outer circle, radius $R$, defines the minimum gaussian distance to the center.}
    \end{center}
    \label{fig:schema3dof}
\end{figure}

Figure \ref{fig:schema3dof} illustrates a typical 3DoF+ scene, with the camera positions in a small spatial
zone compared to the distance to the closest points.

\paragraph{Spherical coordinates}

To make the distance between a gaussian and a camera more explicit, the gaussian
coordinates are now expressed in spherical coordinates $(\rho,\theta,\phi)$.
The origin is set in the area of the cameras centers. With this referential
and the 3DoF+ assumption, the $\rho$ coordinate of any point
approximates the distance between this point and any camera.

Now on, we'll consider a point of spherical coordinates $\rho \mathbf{d}$ is the
unit direction vector defined by:
\begin{align}
    \label{dcoords}
    \mathbf{d} =& (\sin\theta \cos\phi, \sin\theta \sin\phi, \cos\theta)^T
\end{align}

Considering that a camera can point to any direction, we will work on the
$360\degree$ projection. This projection associates each point with its projection
on the sphere of radius $f$.
We will now work in a referential with the camera as center, keeping the
orientation of the world referential.
This referential will be referred to as the local referential.

\begin{align}
    \label{Pcoords}
    \mathbf{P} =& \mathbf{P}_0 + \rho \mathbf{d}
\end{align}
Where $\mathbf{P}_0$ is the world origin in the local referential.

The projection $\mathbf{p}$ of $\mathbf{P}$ on the sphere is:
\begin{align}
    \label{sphereproj}
    \mathbf{p} =&  \frac{\mathbf{P}}{||\mathbf{P}||}
\end{align}

And the derivation of this projection gives (cf details in appendix):
\begin{align}
    \label{projderiv}
    \frac{\partial\mathbf{p}}{\partial \theta} =&  f \frac{\rho}{||\mathbf{P}||} (\cos\theta \cos\phi, \cos\theta  \sin\phi, -\sin\theta)^T + O(\epsilon)\\
    \frac{\partial\mathbf{p}}{\partial \phi}   =&  f \frac{\rho\sin\theta}{||\mathbf{P}||} (-\sin\phi, \cos\phi, 0)^T + O(\epsilon)\nonumber \\
    \frac{\partial\mathbf{p}}{\partial \rho}   =& -f \frac{1}{\rho^2} ((\mathbf{P}_0^T\mathbf{d})\mathbf{d} - \mathbf{P}_0) + o(\epsilon^2) \nonumber \\
    \epsilon =& \frac{||\mathbf{P}_0||}{||\mathbf{P}||} \nonumber
\end{align}

This means that uniform quantization is relevant for $\theta$ and $\phi$, since their impact is bounded by finite values close to 1.
The impact of $\rho$ quantization on the other hand depends on the value of $\rho$ itself. Thus quantizing the coordinate uniformly would be suboptimal.

It is proposed to parameterize $\rho$ as: $\rho = \frac{1}{t}$, yielding:
\begin{align}
    \label{rhoparam}
    \frac{\partial \mathbf{p}}{\partial t}   =& f ((\mathbf{P}_0^T\mathbf{d})\mathbf{d} - \mathbf{P}_0) + o(\epsilon^2)
\end{align}
Which is a good candidate for uniform quantization, since it does not depend on the position of $\mathbf{P}$.

\paragraph{Center vs periphery}
Since the calculations assume the gaussians are far away, compared to the distance between the cameras.
In most cases, this hypothesis is not verified for all gaussians, and quantization scheme can not be used for the whole scene.
A center zone is thus defined, and uses a uniform quantization instead of the proposed model.
Outside of the center zone lies the peripheral zone where the proposed schema is used.

In short:
\begin{enumerate}
    \item In the center, $x,y,z$ are quantized uniformly
    \item In the periphery, $\theta, \phi, 1 / \rho$ are quantized uniformly
\end{enumerate}

The center is defined as the points matching: $\rho < R$ with R being roughly twice the distance
of the cameras from the center of the scene.

\footnote{The simple fact to divide the scene in center vs periphery is already a way to quantize
the coordinates of the central part with a finer step, without changing the quantization step of the
rest of the scene. This optimization can be used on $x,y,z$ coordinate, it is tested in the ablation
study below.}

\section{Experiments}

\paragraph{Test conditions}
We tested the proposed quantization a reconstruction of the Garden scene, from \cite{barron2022mip},
after 30k iterations on the original code of 3DGS.
The PSNR is evaluated on the training views, with the configurations:
\begin{enumerate}
    \item uniform: each x,y,z coordinate is quantized independently, the step depending on the extent of the scene
    \item ours: we use $R$ = 1.5 times the radius of the training cameras positions (bigger values of $R$ did not
    improve the quality of the novel views)
\end{enumerate}

\begin{table}[h!]
    \begin{center}
      \caption{Results on garden scene, mip-nerf dataset}
      \label{tab:table1}
      \begin{tabular}{l|r|r}
        \textbf{bits / coord} & \textbf{uniform} & \textbf{ours}\\
        \hline
        16 & 29.76 & 29.82 \\
        14 & 28.77 & 29.77 \\
        12 & 23.96 & 29.30 \\
      \end{tabular}
    \end{center}
\end{table}

\begin{figure}
    \includegraphics[width=\linewidth]{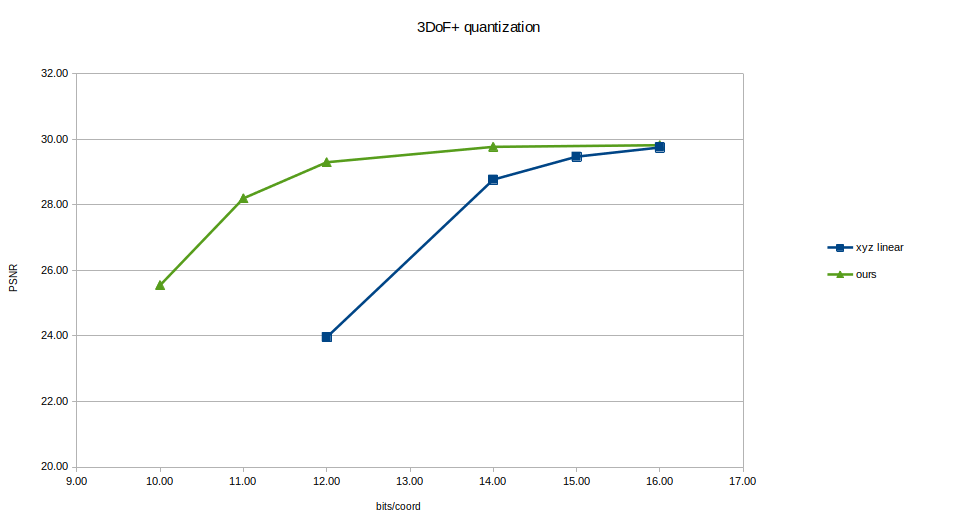}
    \caption{PSNR versus bits/coord}
    \label{fig:psnr}
\end{figure}

The measures listed in table \ref{tab:table1} show a clear improvement
in terms of PSNR when lowering the number of bits per coordinate.

\paragraph{Discussions}
This document proposes an analysis of the impact of quantization noise in terms of projection
on the screen plan. Another impact of quantization noise comes from the use of the position to define in which order
the gaussians are drawn. The analysis of this aspect is left for future work.
The split of the model points in center vs periphery is a new information, which should be added to the
information to be coded. One may argue that the order of the points in the file is an easy way to
encode this information. If the center points are transfered first, only the index of the first
periphery point has to be provided to differentiate the two populations. A more basic way to
transfer this information would be to add one bit per gaussian, which costs 0.33 bit per coordinate.
With this extra cost, the proposed solution keeps better than uniform quantization.

\section{Ablation study}

\begin{figure}
    \includegraphics[width=\linewidth]{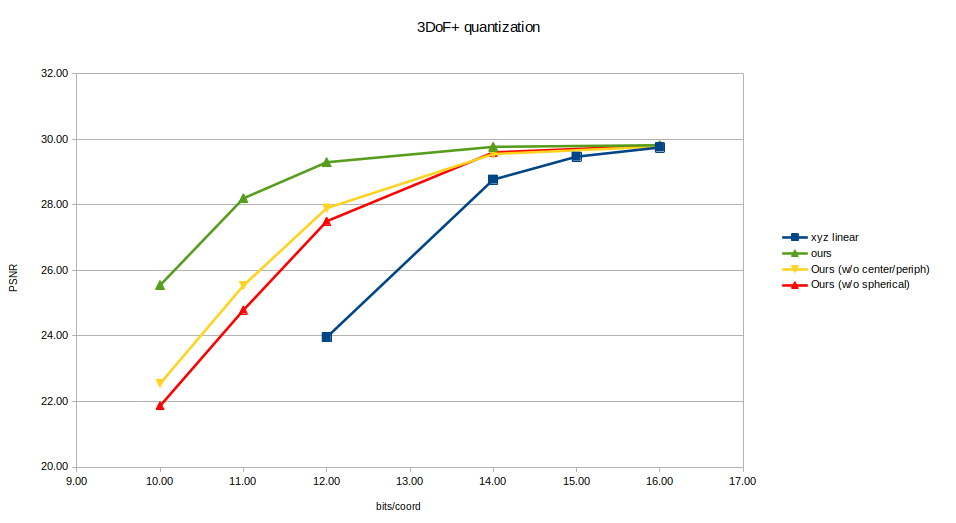}
    \caption{Ablations.}
    \label{fig:ablations}
\end{figure}

The proposed quantization scheme includes the use of spherical coordinates and the
split of the scene in two parts: center vs periphery.
Spherical coordinates, including the inversion of $\rho$, bring some value by 
giving more precision to points close to the center. The added value of this part
is illustrated by \textit{"w/o center/periphery"}, where $1/\rho,\theta,\phi$ are
quantize independently of the gaussian position.
The split in center vs periphery is another way to enable finer precision
on the center, without sacrifying the precision of the periphery. This part
is illustrated by \textit{"w/o spherical"}, where $x,y,z$ are quantized, with two sets
of bounds: $[-R,R]$ for the center gaussians, and the extent of the whole scene for the periphery.
The figure \ref{fig:ablations} shows the PSNR reached at different bits/coord values.

\section{Conclusions}
This article proposes a simple parameterization of spatial coordinates, which minimizes the
projection error due to the positions quantization, compared to a standard uniform quantization.
This straightforward technique does not interfere with 3DGS training and is compatible with many
other compression algorithms.
Though exposed in a simple 3DoF+ context, this technique can be adapted to many 3D scenes
to avoid storing and transmitting more information than needed for zones that will be viewed far away
at render time.

\pagebreak
\begin{appendices}

\section{Partial derivations}
This section details the partial derivatives exposed in the document.

Unit vector of the spherical coordinates:
\begin{align}
    \label{projderivd}
    \mathbf{d} =& (\sin\theta \cos\phi, \sin\theta \sin\phi, \cos\theta)^T\nonumber \\
    \frac{\partial \mathbf{d}}{\partial \theta} =& (\cos\theta \cos\phi, \cos\theta \sin\phi, -\sin\theta)^T \nonumber \\
    \mathbf{d}^T\frac{\partial \mathbf{d}}{\partial \theta} =& 0 \nonumber \\
    \frac{\partial \mathbf{d}}{\partial \phi}   =& (- \sin\theta \sin\phi, \sin\theta \cos\phi, 0)^T \nonumber \\
    \mathbf{d}^T\frac{\partial \mathbf{d}}{\partial \phi} =& 0
\end{align}

Point in the camera referential:
\begin{align}
    \label{projderivp}
    \mathbf{P} =&  \mathbf{P}_0 + \rho \mathbf{d} \nonumber \\
    \frac{\partial \mathbf{P}}{\partial \rho} =& \mathbf{d} \nonumber \\
    \frac{\partial \mathbf{P}}{\partial \mathbf{d}}   =& \rho  \mathbf{I}
\end{align}
With $\mathbf{I}$ the $3\times3$ identity matrix.

Projection of the point on the unit sphere:
\begin{align}
    \label{dpdP}
    ||\mathbf{P}|| =& \sqrt{p_x^2 + p_y^2 + p_z^2} \nonumber \\
    \frac{\partial ||\mathbf{P}||}{\partial \mathbf{P}} =& (p_x / \sqrt{p_x^2 + p_y^2 + p_z^2}, p_y / \sqrt{p_x^2 + p_y^2 + p_z^2}, p_z / \sqrt{p_x^2 + p_y^2 + p_z^2})\nonumber \\
                                      =& \frac{1}{||\mathbf{P}||} \mathbf{P}^T\nonumber \\
    \mathbf{p} =& \frac{1}{||\mathbf{P}||} \mathbf{P}\nonumber \\
    \frac{\partial \mathbf{p}}{\partial \mathbf{P}} =& \frac{1}{||\mathbf{P}||}  \mathbf{I} + \mathbf{P} (\frac{-1}{||\mathbf{P}||^2} \frac{\partial ||\mathbf{P}||}{\partial \mathbf{P}})\nonumber \\
                                                    =& \frac{1}{||\mathbf{P}||}  \mathbf{I} + \mathbf{P} (\frac{-1}{||\mathbf{P}||^2} \frac{1}{||\mathbf{P}||} \mathbf{P}^T)\nonumber \\
                                                    =& \frac{1}{||\mathbf{P}||}  \mathbf{I} - \frac{1}{||\mathbf{P}||^3} \mathbf{P} \mathbf{P}^T
\end{align}
Derivation of the projection by angles:
\begin{align}
    \label{dppda}
    \frac{\partial \mathbf{p}}{\partial \theta} =& (\frac{1}{||\mathbf{P}||}  \mathbf{I} - \frac{1}{||\mathbf{P}||^3} \mathbf{P} \mathbf{P}^T) \frac{\partial \mathbf{P}}{\partial \mathbf{d}} \frac{\partial \mathbf{d}}{\partial \mathbf{\theta}}\nonumber \\
    =& (\frac{1}{||\mathbf{P}||}  \mathbf{I} - \frac{1}{||\mathbf{P}||^3} \mathbf{P} \mathbf{P}^T) \rho \frac{\partial \mathbf{d}}{\partial \mathbf{\theta}}\nonumber \\
    =& \frac{1}{||\mathbf{P}||} \rho \frac{\partial \mathbf{d}}{\partial \theta} - \frac{1}{||\mathbf{P}||^3} \mathbf{P} \mathbf{P}^T (\rho \frac{\partial \mathbf{d}}{\partial \theta})\nonumber \\
    =& \frac{\rho}{||\mathbf{P}||} (\frac{\partial \mathbf{d}}{\partial \theta} - \frac{1}{||\mathbf{P}||}(\mathbf{P}_0^T\frac{\partial \mathbf{d}}{\partial \theta})\mathbf{p})\nonumber \\
    =& \frac{\rho}{||\mathbf{P}||} \frac{\partial \mathbf{d}}{\partial \theta} + O(\epsilon)\nonumber \\
    =& \frac{\rho}{||\mathbf{P}||} (\cos\theta \cos\phi, \cos\theta \sin\phi, -\sin\theta) + O(\epsilon)\nonumber \\
    \frac{\partial \mathbf{p}}{\partial \phi} =& (\frac{1}{||\mathbf{P}||}  \mathbf{I} - \frac{1}{||\mathbf{P}||^3} \mathbf{P} \mathbf{P}^T) \frac{\partial \mathbf{P}}{\partial \mathbf{d}} \frac{\partial \mathbf{d}}{\partial \mathbf{\phi}}\nonumber \\
    =& \frac{1}{||\mathbf{P}||} \rho \frac{\partial \mathbf{d}}{\partial \phi} - \frac{1}{||\mathbf{P}||^3} \mathbf{P} \mathbf{P}^T(\rho \frac{\partial \mathbf{d}}{\partial \phi})\nonumber \\
    =& \frac{\rho}{||\mathbf{P}||} (\frac{\partial \mathbf{d}}{\partial \phi} - \frac{1}{||\mathbf{P}||}(\mathbf{P}_0^T\frac{\partial \mathbf{d}}{\partial \phi})\mathbf{p})\nonumber \\
    =& \frac{\rho}{||\mathbf{P}||} \frac{\partial \mathbf{d}}{\partial \phi} + O(\epsilon)\nonumber \\
    =& \frac{\rho \sin\theta}{||\mathbf{P}||} (- \sin\phi, \cos\phi, 0) + O(\epsilon)
\end{align}
Derivation of the projection by $\rho$:
\begin{align}
    \label{dppdr}
    \frac{\partial \mathbf{p}}{\partial \rho} =& (\frac{1}{||\mathbf{P}||}  \mathbf{I} - \frac{1}{||\mathbf{P}||^3} \mathbf{P} \mathbf{P}^T) \frac{\partial \mathbf{P}}{\partial \mathbf{\rho}}\nonumber \\
    =& \frac{1}{||\mathbf{P}||} \mathbf{d} - \frac{1}{||\mathbf{P}||^3} \mathbf{P} \mathbf{P}^T\mathbf{d} \nonumber \\
    =& \frac{1}{||\mathbf{P}||} \mathbf{d} - \frac{1}{||\mathbf{P}||^3} ( \mathbf{P}_0 + \rho \mathbf{d}) (\rho + \mathbf{P}_0^T\mathbf{d})    \nonumber \\
    =& \frac{1}{||\mathbf{P}||^3} (||\mathbf{P}||^2 \mathbf{d} - \rho  \mathbf{P}_0 - (\mathbf{P}_0^T\mathbf{d})\mathbf{P}_0 - \rho^2 \mathbf{d} - \rho (\mathbf{P}_0^T\mathbf{d}) \mathbf{d})   \nonumber \\
    =& \frac{1}{||\mathbf{P}||^3} ((\rho^2 + 2 \rho \mathbf{P}_0^T\mathbf{d} + || \mathbf{P}_0||^2) \mathbf{d} - \rho  \mathbf{P}_0 - (\mathbf{P}_0^T\mathbf{d})\mathbf{P}_0 - \rho^2 \mathbf{d} - \rho (\mathbf{P}_0^T\mathbf{d}) \mathbf{d})   \nonumber \\
    =& \frac{1}{||\mathbf{P}||^3} ((\rho \mathbf{P}_0^T\mathbf{d} + || \mathbf{P}_0||^2) \mathbf{d} - (\rho + \mathbf{P}_0^T\mathbf{d})  \mathbf{P}_0)   \nonumber \\
    =& \frac{\rho}{||\mathbf{P}||} \frac{1}{||\mathbf{P}||^2} ((\mathbf{P}_0^T\mathbf{d})\mathbf{d} -  \mathbf{P}_0) + \frac{1}{||\mathbf{P}||} (\frac{|| \mathbf{P}_0||^2}{||\mathbf{P}||^2} \mathbf{d} - \frac{\mathbf{P}_0^T\mathbf{d}}{||\mathbf{P}||} \frac{ \mathbf{P}_0}{||\mathbf{P}||})   \nonumber \\
    =& \frac{\rho}{||\mathbf{P}||} \frac{1}{||\mathbf{P}||^2} ((\mathbf{P}_0^T\mathbf{d})\mathbf{d} -  \mathbf{P}_0) +  O(\epsilon^2)  \nonumber \\
    =& \frac{1}{||\mathbf{P}||^2} ((\mathbf{P}_0^T\mathbf{d})\mathbf{d} -  \mathbf{P}_0) +  O(\epsilon^2)
\end{align}
Please note that in these equations, $\mathbf{p}$ is the projection on the unit sphere, as in the paper it's the projection on the sphere of radius f.

\end{appendices}

\bibliography{quantization3DoF}
\bibliographystyle{apalike}

\end{document}